\documentclass[10pt,twocolumn,letterpaper]{article}

\usepackage{cvpr}
\usepackage{times}
\usepackage{epsfig}
\usepackage{graphicx}
\usepackage{amsmath}
\usepackage{multirow}
\usepackage{amssymb}
\usepackage{tabularx}

\usepackage[pagebackref=true,breaklinks=true,letterpaper=true,colorlinks,bookmarks=false]{hyperref}

\cvprfinalcopy 

\def\cvprPaperID{5316} 

\ifcvprfinal\fi
\begin{document}

\title{Two-stream Convolutional Networks for Multi-frame Face Anti-spoofing}


\author{Zhuoyi Zhang\textsuperscript{1,2}, Cheng Jiang\textsuperscript{2}, Xiya Zhong\textsuperscript{2}, Chang Song\textsuperscript{1,2}, Yifeng Zhang\textsuperscript{1}\\
\textsuperscript{1}School of Signal and Information Processing, Southeast University\\
Nanjing, China\\
\textsuperscript{2}SenseTime Group Limited \\
Shanghai, China\\
{\tt\small \{zzy96,yfz\}@seu.edu.cn, \{jiangcheng,zhongxiya,songchang\}@sensetime.com}}

\maketitle

\begin{abstract}
Face anti-spoofing is an important task to protect the security of face recognition. Most of previous work either struggle to capture discriminative and generalizable feature or rely on auxiliary information which is unavailable for most of industrial product. Inspired by the video classification work, we propose an efficient two-stream model to capture the key differences between live and spoof faces, which takes multi-frames and RGB difference as input respectively. Feature pyramid modules with two opposite fusion directions and pyramid pooling modules are applied to enhance feature representation. We evaluate the proposed method on the datasets of Siw, Oulu-NPU, CASIA-MFSD and Replay-Attack. The results show that our model achieves the state-of-the-art results on most of datasets' protocol with much less parameter size.
\end{abstract}

\section{Introduction}

Face authentication is being increasingly used in our daily life for tasks such as phone unlocking, access authentication and face-payment owing to its convenience and high-efficiency. While face authentication systems are vulnerable to spoofing attacks with printed photos, relayed videos and forged masks without protective measures. Therefore, face anti-spoofing is vital to the security of face recognition systems, which can defend the above mentioned attacks efficiently~\cite{Di2015Face,Unar2014A}.
\begin{figure}[t]
\begin{center}
\includegraphics[width=0.9\linewidth]{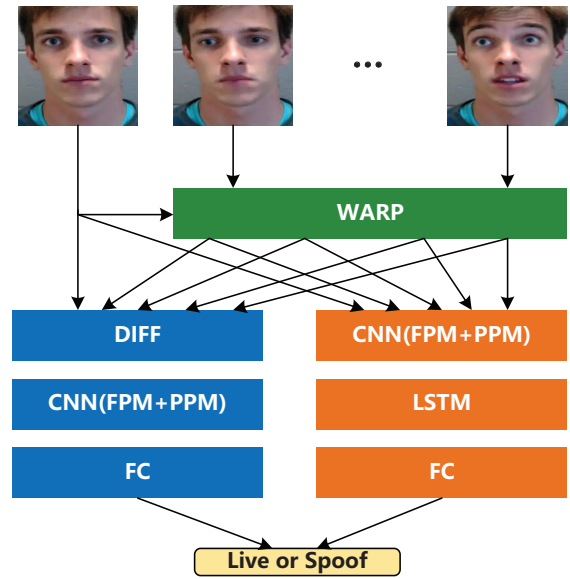}
\end{center}
   \caption{The pipeline of our proposed face anti-spoofing network. The origin multiple frames are first warped to align according to the initial frame. Left is the RGB difference branch which takes stacked RGB difference image as input. It has a feature pyramid feature module and a pyramid pooling module to extract robust temporal features. Right is the multi-frame branch consists of a CNN-LSTM model, an opposite feature pyramid module(FPM) and a pyramid pooling module(PPM). The final prediction is determined by the combination of two branches' prediction score.}
\label{fig1}
\end{figure}

The core problem for face anti-spoofing is how to capture discriminative features among genuine faces and attackers. Its purpose is to predict whether a face in a video belongs to a genuine person, which can be deemed as a binary classification task. While typical classification model can not meet requirements of face anti-spoofing since features they need are varied in granularity. Besides, speed is an important factor in industry, thus lightness for model is considered to be a significant attribute to evaluate.

Previous methods have made some progress in recent years. A typical method is tacking face anti-spoofing task from the aspect of space. In our view, the differences in the dimension of spatial lie in two aspects: texture and distribution pattern. Most prior works focus on texture, regard it as a noise extraction task. While in the condition of low quality face images or videos, the texture feature is indistinct and hard to distinguish, especially when we utilize a deep model, who will lose much detail information because of down-sampling. Some methods try to divide image into several patches to focus more on its texture, but it breaks the distribution of origin images and increases computational complexity. The distribution pattern includes perspective deformation and spoofing artifacts caused by screen sloping and distortion when warping a print paper or photo. It can only be observed through the full face image.

Another perspective is capturing differences in sequential. They consider the motion patterns contained in genuine and spoofing videos are different. For a genuine face, its motion cues are mainly expression changes such as eye blinking, mouth movement. While the temporal information in print or photo attacks are mainly caused by hand-trembling or material reflection, and there are also expression changes in replay attack. The motion cues are very valuable to distinguish valid access with attempted attacks, and also the relative motion between the face region and the background can be helpful. This kind of methods usually need multiple frames, and are expected to have better performance than methods using single frame. While unlike normal video classification task, we need more detailed temporal features because of the subtle movement of face verification video. Besides, for replay attack, its temporal information is similar with live faces except for some perspective changes due to the depth of screen, and need more fine-grained temporal features.

There are some multi-modal based methods show superiority for defending replay attack and print attack, which benefit of multi-dimension data, such as depth images or IR images. In practical these kind of data is hard to obtain considering cost and application condition. For the sake of accessibility and costs, we focus on the ubiquitous 2D-camera case, available in almost all mobile devices and easy to acquire. While a vital problems for anti-spoofing task with RGB images is over-fitting since origin frame contain full information of face appearance, which is valuable for face recognition while redundant for face anti-spoofing. Some works try to break the image distribution and then recover it to ignore face appearance, but they also lose the distortion and perspective information that may helpful to recognize hack faces. Some~\cite{Jourabloo2018Face} try to regard hack images as live faces plus noises, and try to model noises and then classify by them. But the noises is hard to model and has high dependence on the image quality.

In this paper, we form a light-weight two-stream model to handle the above problems. Our model consists of two branches: multi-frame branch and RGB difference branch. The former utilize multiple RGB images as inputs and pay attention to detailed spatial information by a feature pyramid module and a pyramid pooling module, and use LSTMs to capture temporal features. The later take RGB difference images as inputs. By applying a feature pyramid module with the opposite direction and a pyramid pooling module, it can capture fine-grained temporal feature and avoid over-fitting. The feature enhancement modules make the two branches become more powerful respectively and complementary to each other. By combining the two branches with much less parameter size, it reaches a competitive accuracy compared with state-of-the-art methods without using auxiliary data or heavy model under most evaluation metrics.

Our contributions can be summarized as follows:
\begin{itemize}
\item We design a novel CNN-LSTM model to capture discriminative spatial-temporal features.
\item We form a branch take RGB difference as inputs to learn key differences between genuine face and attacks and decrease over-fitting.
\item We propose to utilize feature pyramid modules with different fusion direction and pyramid pooling modules in two-stream network to obtain robust spatial and temporal features for both two branches.
\item We conduct extensive experimental analysis on the datasets of Siw, Oulu-NPU, CASIA-MFSD and Replay-Attack. The results show that our proposed network can achieve better performance compared with other state-of-the-art methods without auxiliary data and with much less parameter size on most of datasets' protocol.
\end{itemize}

\section{Related Work}

\textbf{Face anti-spoofing used texture information.}
The texture-based analysis exploited the fact that real face contains different texture and illumination pattern as compared to a plastic or LCD surface. Hand-crafted features are utilized to capture texture features, such as LBP~\cite{Pereira2012LBP}, HoG~\cite{Komulainen2014Context} and SIFT~\cite{Patel2016Secure}, and then classified by a SVM. Wen et al.~\cite{Di2015Face} considered four types of surface deformations such as specular reflection, blurriness features, chromatic moment and color diversity to generate the feature vector and used SVM classifier to classify the output feature into live or not. The above methods have several drawbacks including the need to utilize hand designed features and the limited performance. Since convolutional neural networks have been proved to be effective in classification task, some CNN based methods try to extract more robust spatial features to distinguish real from spoofed faces. Li et.al~\cite{Li2016An} utilized a VGGNet~\cite{Simonyan2014Very} pre-trained on ImageNet~\cite{Russakovsky2015ImageNet} to capture deep robust features, and used a SVM to classify. Nagpal et. al~\cite{Nagpal2018A} evaluated the performance of utilizing several popular CNN structure including ResNet~\cite{He2016Deep} and GoogLeNet~\cite{Szegedy2014Going} in face anti-spoofing. They also validate the influences of choosing different hyper parameters for training. Tu et. al~\cite{Tu2019Deep} proposed a CNN framework using sparsely labeled data from the target domain to learn features that are invariant across domains for face anti-spoofing, which improved the generalization ability across different kinds of databases.

\textbf{Face anti-spoofing used spatial-temporal information.}
The motion based face anti-spoofing is also investigated by several researchers by exploiting the fact that the most of face attacks happen with the use of stills and thus, its motion pattern can be used to differentiate a live subject. Wang et. al~\cite{Wang2018Exploiting} proposed a depth supervised structure with OFF block and ConvGRU module to uncover facial depths and their unique motion patterns from temporal information of monocular frame sequences. Liu et. al~\cite{Liu2018Learning} proposed a novel CNN-RNN structure for end to-end learning the depth map and rPPG signal. They improved model generalization by training with supervision of depth map and rPPG signal. Li et. al~\cite{Li2018Learning} utilized a 3D CNN network which take both spatial and temporal information into consideration. They decrease over-fitting problem by a specifically designed data augmentation method and a generalization regularization. Tu et.al~\cite{Tu2019Enhance} leveraged LSTM with the extracted features as inputs to capture the temporal dynamics in videos. To ensure the fine-grained motions more easily to be perceived in the training process, the Eulerian motion magnification is used as the preprocessing to enhance the facial expressions exhibited by individuals, and the attention mechanism is embedded in LSTM to ensure the model learn to focus selectively on the dynamic frames across the video clips. Yang et. al~\cite{Yang2019Face} exploited a novel spatial-temporal anti-spoofing network, which can automatically attend to discriminative regions, and it makes analyzing the behaviors of the network possible. They conducted extensive experiments and show that the proposed model can distinguish spoof faces by extracting features from a variety of regions to seek out subtle evidences such as borders, moire patterns, reflection artifacts, etc.

\textbf{Two-stream neural network.}
Essentially face anti-spoofing is a classification task, and it can be regarded as video classification when input a video. Therefore, many ideas of video classification can be applied in our task. Simonyan et. al~\cite{Simonyan2014Two} first used a two-stream neural network for action recognition in video which incorporates spatial and temporal networks by single frame and multi-frame optical flow inputs, and obtained competitive performance. TSN~\cite{Wang2016Temporal} utilized two-stream ConvNets to classify videos and investigated the influence of utilizing different input modality including RGB images, RGB difference, optical flow fields, and warped optical flow fields. They observed that the combination of RGB images and RGB differences can boost the recognition performance since they encode complementary information, and RGB difference can serve as a low-quality, high-speed alternative for motion representations. Sun et. al~\cite{Sun2017Optical} introduced a novel compact motion representation for video action recognition, named Optical Flow guided Feature (OFF), which enables the network to distill temporal information through a fast and robust approach. The OFF which derived from the definition of optical flow provides theoretical support for using the difference between two frames. By directly calculating pixel-wise spatiotemporal gradients of the deep feature maps, the OFF could be embedded in any existing CNN based video action recognition framework with only a light additional cost.

\section{Our Approach}
In the following, we describe the proposed method for face anti-spoofing. The pipeline of our method is shown in Fig.~\ref{fig1}. Our algorithm consists of a multi-frame branch and a RGB difference branch. The multi-frame branch is CNN-LSTM structure to extract spatial and temporal features, and RGB difference branch is a classification model to extract temporal features and decrease over-fitting. We first present the preprocess of video data, and then the multi-frame model, finally describe the RGB difference model.
\subsection{Preprocess}
In face authentication, what we are given is a video clip contains a certain face. After extracting frames from a video, there may exist much background information which interfere our inference. We first use a face detector to recognize key points for face images. For the common methods of face anti-spoofing, the face region is then cropped to fed into a classification model. While we observed that if face move dynamically, it is hard to distinguish the key temporal features since both genuine and fake faces can generate sweeping movements. If we align faces to eliminate the variations in the spatial appearances first, we can focus on the subtle motion cues in facial expression changes. We normalise the face geometry according to the initial frame in a video by choosing anchor points, whose position are fixed even with face expression changes. For a point $(x,y)$, its registered value is
\begin{equation}
f(x,y)=g(\frac{W_{1}x+W_{2}y+W_{3}}{W_{7}x+W_{8}y+W_{9}},\frac{W_{4}x+W_{5}y+W_{6}}{W_{7}x+W_{8}y+W_{9}})
\end{equation}
where $g$ is the origin image and $W$ is the weight decided by anchor points.

After registration, we compute a crop box according to the first frame's key points, and all the following frames are cropped into a face region according to this bounding box. Then a fixed number of key frames are located roughly by computing difference deviation. All frames in a video are divided into $k$ stacks, and the first frame is defined as the onset frame. In the first stack we find a frame which have the greatest changes from the onset frame, i.e. have the largest difference deviation, and regard this frame as the new onset frame. We can collect $k+1$ frames (including the first frame) which changes dynamically to capture more temporal information.
\begin{figure}[t]
\begin{center}
\includegraphics[width=0.9\linewidth]{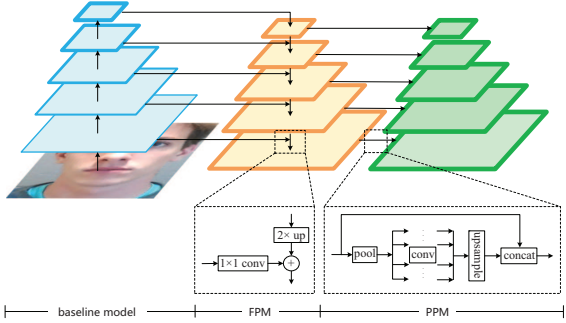}
\end{center}
   \caption{The spatial feature extractor based on MobileNetV3-Small model in multi-frame branch. The blue blocks are origin feature maps at different levels, and the yellow parts are outputs of the feature pyramid module. Green blocks stand for the final spatial feature maps processed by the pyramid pooling module. These arrows indicate feature spread direction.}
\label{fig2}
\end{figure}
\subsection{Multi-frame branch}
The branch of multi-frame takes multiple frames as inputs, which are selected by our sampling logic from a video. For each frame, we utilize a MobileNetV3-Small~\cite{Howard2019Searching} neural network as spatial feature extractor. As discussed above, detailed feature(texture) and global feature(distribution) are both important for defending spoofing faces, therefore we add a feature pyramid module and a pyramid pooling module to capture hierarchical and multi-granularity spatial features for enhancing representations for key features in face anti-spoofing, as shown in Fig.~\ref{fig2}.

In our view, the output feature map in the origin network may ignore some important details because of down-sampling. As a result, we spread robust deep features from shallow layers to deep layers which may encode rich detail information by applying a feature pyramid module~\cite{Lin2016Feature}. In our network, we up-sample the coarser-resolution feature map by a factor of $2$ and merge it with the corresponding bottom-up map by element-wise addition, where $1\times1$ convolution is applied to adjust the number of channel. We perform this operation once after each down-sampling. Since there are five down-sample layers in our network, we obtain five spatial feature representations for each frame at different levels.

After obtaining hierarchical features, we need more context information to perceive global distribution for a frame. For each feature map, a pyramid pooling module~\cite{Zhao2017Pyramid} is applied to harvest different sub-region representations by average pooling operation with varying-size pooling kernels in a few strides to get four bins with size of $1\times1$, $2\times2$, $3\times3$ and $6\times6$ respectively, and then followed by up-sampling and concatenation layers to form the final feature representation, which carries multi-granularity information including both local and global context features.

\begin{figure}[t]
\begin{center}
\includegraphics[width=0.9\linewidth]{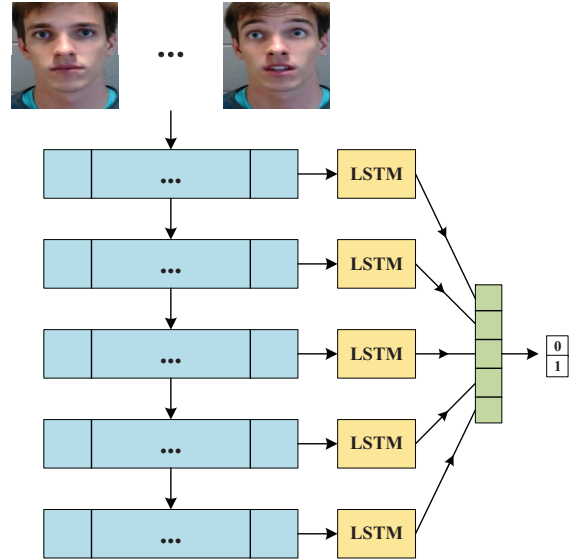}
\end{center}
   \caption{The temporal feature extractor in multi-frame branch. Features obtained from the same layers are fed into a Bi-LSTM to encode temporal features in a certain receptive field. All temporal features are then concatenated to compose a final feature. A fully-connected layer is applied to predict scores for distinguish live and spoofing faces.}
\label{fig3}
\end{figure}
For one frame, we have obtained its robust and fine-grained spatial features at five levels. We utilize a CNN-LSTM structure to concatenate multiple frames' feature at each level as a sequence. As shown in Fig.~\ref{fig3}, feature maps at the same levels are concatenated to fed into a Bi-LSTM respectively to model temporal correlation across multiple frames at different levels. The outputs of them are concatenated again into a final layer which encodes hierarchical spatial-temporal features. Followed by a fully-connected layer and a softmax activation function, the prediction result of face anti-spoofing for multi-frame branch is determined.
\subsection{RGB difference branch}
The multiple frames extracted from a video represent its spatial and motion cues, but sometimes less is more. For face anti-spoofing too much information means more risks of over-fitting. Inspired by two-stream methods in video classification~\cite{Simonyan2014Two,Wang2016Temporal}, we also design another branch which take RGB differences as inputs. For the most conditions of face authentication, the video clips are relatively short and have a fix background, RGB differences between two consecutive frames can help us to focus on motion information generated in faces efficiently. Since we have applied registration in preprocess, the RGB differences contain motion cues without still spatial distribution information, which can avoid over-fitting into human faces or background.

\begin{figure}[t]
\begin{center}
\includegraphics[width=0.9\linewidth]{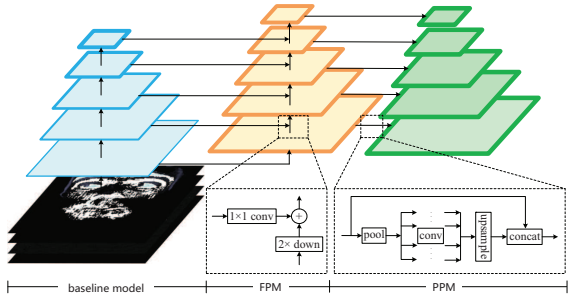}
\end{center}
   \caption{The structure of our proposed RGB difference branch, which based on MobileNetV3-Small network. The blue, yellow and green blocks represent for origin features, FPM features and PPM features respectively. The feature spread direction is opposite to the multi-frame branch. The five PPM features are directly fed in a fully-connected layer without LSTM to obtain fine-grained features.}
\label{fig4}
\end{figure}
After concatenating multiple RGB difference images, we fed it into another modified MobileNetV3-Small neural network, whose main difference with the multi-frame RGB branch is the feature pyramid module. Unlike RGB frames branch whose features from deeper layers are robust since it encode abundant information, there are less information among RGB difference images, and we need more shallow features because detailed features are more important than its semantic feature for motion cues. Therefore, we make a modification for feature pyramid module, which spread information from shallow layers to deeper layers. As shown in Fig.~\ref{fig4}, after each layer's down-sampling, we down-sample the larger-resolution feature map by max pooling and merge it with the corresponding top-down map by element-wise addition, where $1\times1$ convolution is applied to adjust the number of channel.

Followed by the feature pyramid module is a pyramid pooling module to capture multi-granularity features. Since RGB difference inputs have encoded motion information itself, these features are concatenated directly without LSTMs, and then fed into a fully-connected layer and a softmax activation function to predict scores for live faces and spoof faces.
\subsection{Models training}
Benefit of different input modal and feature fusion direction, this two branches show great complementarity. We train them independently. For multi-frame branch, we exploit a baseline model trained on the ImageNet as initialization, and then fine tune the whole parameters on training set. We also utilize a pretrained RGB model to initialize the RGB difference network, which has two differences in training process. The first is we do not need to standardize inputs as multi-frame branch does, since it already has normal distribution around zero. Second is we need to modify the weights of first convolution layer of pretrained models to handle the input of RGB difference. We average the weights across the RGB channels and replicate this average by the channel number of RGB difference network input. After training, the final prediction is determined by the sum value of these two branches' predict scores.

\section{Experiment}
\subsection{Datasets and Metrics}
We evaluate the proposed model on four public face anti-spoofing databases, including Replay-Attack~\cite{Chingovska2012On}, CASIA-MFSD~\cite{Zhang2012A}, Oulu-NPU~\cite{Boulkenafet2017OULU} and SiW~\cite{Liu2018Learning}.
The Replay-Attack Database~\cite{Chingovska2012On} for face spoofing consists of 1300 video clips of photo and video attack attempts to 50 clients, under different lighting conditions, which includes print and replay attack. CASIA-MFSD~\cite{Zhang2012A} contains 50 genuine subjects, and fake faces are made from the high quality records of the genuine faces. Three fake face attacks are implemented, which include warped photo attack, cut photo attack and video attack, and each subject contains 12 videos (3 genuine and 9 fake), and the final database contains 600 video clips. The Oulu-NPU~\cite{Boulkenafet2017OULU} face presentation attack detection database consists of 4950 real access and attack videos. The presentation attack types considered in the OULU-NPU database are print and video-replay generated by different printer and display devices. SiW~\cite{Liu2018Learning} provides live and spoof videos from 165 subjects. For each subject, we have 8 live and up to 20 spoof videos, in total 4,478 videos. All videos are in 30 fps, about 15 second length, and 1080P HD resolution. The live videos are collected in four sessions with variations of distance, pose, illumination and expression. The spoof videos are collected with several attacks such as printed paper and replay.

We employ the ACER and HTER to evaluate performance~\cite{ISO}, where ACER is the mean value of the Attack Presentation Classification Error Rate (APCER) and the Bona Fide Presentation Classification Error Rate (BPCER), and the HTER is half of the sum of the False Rejection Rate (FRR) and the False Acceptance Rate (FAR).

\subsection{Implementation details}
Our model is implemented with PyTorch framework. After face detection for each frame, we resize each face region to the fixed size of $224\times224$. We use stochastic gradient descent algorithm to optimize our proposed network with initial learning rate of 0.01. The batch size of the two branch are 8 and 16 respectively and the number of frames extracted from a video is 10 for intra testing and 5 for cross testing, i. e $k$ is $9$ and $4$ respectively, and for longer video clips we can sample more frames. During training, spatial and temporal data augment methods are used, which are random clip and crop for spatial augment and resample frames for temporal augment.

The parameter size of the whole two-stream model is only 1.73 M, since we remove the last fully-connected layers of the original MobileNetV3-Small. Its lightness and effectiveness allow it to be implemented widely in practical, such as mobile devices, and bring better user experience.
\subsection{Intra Testing}
The intra testing denotes that evaluating our methods with testing data which belongs to the same database with training data. We carried intra testing on Onlu-NPU~\cite{Boulkenafet2017OULU} and SiW~\cite{Liu2018Learning} databases. The ACER is used to report quantitative results in SiW database. There are three protocols defined by SiW: Protocol 1 is designed to evaluate the generalization of the face PAD methods under different face poses and expression. Protocol 2 evaluates the generalization capability on cross medium of the same spoof type. Protocol 3 is designed to evaluate the performance on unknown PA. We follow their rules to train and test. For evaluating, we compare our algorithm with recent excellent methods including Auxiliary~\cite{Liu2018Learning}, OFFB~\cite{Wang2018Exploiting} and STASN~\cite{Yang2019Face}, which are all CNN-LSTM like structure. Among these methods, we have the lightest model and do not need any auxiliary data. Auxiliary utilized depth map and rPPG signal as auxiliary supervision. OFFB used depth map as label to train. STASN introduced a heavy model to train and infer. Table~\ref{table1} shows the comparative results of performance in SiW database. For the consideration of protocol 1 and 2, our method performs best among these state-of-the-art methods, show its great generalization under different face expressions and mediums. The disadvantage of our model is that we can not handle unknown presentation attacks well because we lack of auxiliary data to capture differences in different perspective.
\begin{table}[t]
\begin{center}
\begin{tabular}{|c|c|c|}
\hline
Prot.&Method&ACER($\%$) \\
\hline
\multirow{4}*{1}
&Auxiliary~\cite{Liu2018Learning}&3.58 \\
\cline{2-3}
&OFFB~\cite{Wang2018Exploiting}&0.73 \\
\cline{2-3}
&STASN~\cite{Yang2019Face}&1.0 \\
\cline{2-3}
&Ours&\textbf{0.26} \\
\hline
\multirow{4}*{2}
&Auxiliary~\cite{Liu2018Learning}&0.57$\pm$0.69 \\
\cline{2-3}
&OFFB~\cite{Wang2018Exploiting}&\textbf{0.15$\pm$0.14} \\
\cline{2-3}
&STASN~\cite{Yang2019Face}&0.28$\pm$0.05 \\
\cline{2-3}
&Ours&\textbf{0.15$\pm$0.21} \\
\hline
\multirow{4}*{3}
&Auxiliary~\cite{Liu2018Learning}&8.31$\pm$3.80 \\
\cline{2-3}
&OFFB~\cite{Wang2018Exploiting}&\textbf{3.10$\pm$0.81} \\
\cline{2-3}
&STASN~\cite{Yang2019Face}&12.1$\pm$1.50 \\
\cline{2-3}
&Ours&14.76$\pm$3.43 \\
\hline
\end{tabular}
\end{center}
\caption{Comparative results of intra test on SiW database.}
\label{table1}
\end{table}

For evaluating on Oulu-NPU dataset, it considers four protocols: Protocol 1 is designed to evaluate the generalization under previously unseen environmental conditions, namely illumination and background scene. Protocol 2 aims to evaluate the effect of attacks created with different printers or displays devices. Protocol 3 is used to study the effect of the input camera variation. Protocol 4 considers generalization of face PAD methods across previously unseen environmental conditions, attacks and input sensors. By using the required fashion to apply, we show the results in Table~\ref{table2} compared with Auxiliary~\cite{Liu2018Learning}, STASN~\cite{Yang2019Face}, OFFB~\cite{Wang2018Exploiting}, MixedFASNet~\cite{Boulkenafet2017A} and GRADIANT~\cite{Boulkenafet2017A}. Our proposed method achieves two best and two second ACER scores in four protocols among all state-of-the-art methods, indicates that light model without auxiliary supervision can achieve competitive performance in intra testing.
\begin{table}[t]
\begin{center}
\resizebox{8.3cm}{4.0cm}{
\begin{tabular}{|c|c|c|c|c|}
\hline
Prot.&Method&APCER($\%$)&BPCER($\%$)&ACER($\%$) \\
\hline
\multirow{5}*{1}
&Auxiliary~\cite{Liu2018Learning}&1.6&1.6&1.6 \\
\cline{2-5}
&GRADIANT~\cite{Boulkenafet2017A}&1.3&12.5&6.9 \\
\cline{2-5}
&OFFB~\cite{Wang2018Exploiting}&2.5&\textbf{0}&1.3 \\
\cline{2-5}
&STASN~\cite{Yang2019Face}&1.2&2.5&1.9 \\
\cline{2-5}
&Ours&\textbf{0.9}&1.5&\textbf{1.2} \\
\hline
\multirow{6}*{2}
&Auxiliary~\cite{Liu2018Learning}&2.7&2.7&2.7 \\
\cline{2-5}
&GRADIANT~\cite{Boulkenafet2017A}&3.1&1.9&2.5 \\
\cline{2-5}
&MixedFASNet~\cite{Boulkenafet2017A}&9.7&2.5&6.1 \\
\cline{2-5}
&OFFB~\cite{Wang2018Exploiting}&\textbf{1.7}&2.0&\textbf{1.9} \\
\cline{2-5}
&STASN~\cite{Yang2019Face}&4.2&\textbf{0.3}&2.2 \\
\cline{2-5}
&Ours&3.3&0.9&2.1 \\
\hline
\multirow{6}*{3}
&Auxiliary~\cite{Liu2018Learning}&2.7$\pm$1.3&3.1$\pm$1.7&2.9$\pm$1.5 \\
\cline{2-5}
&GRADIANT~\cite{Boulkenafet2017A}&\textbf{2.6}$\pm$3.9&5.0$\pm$5.3&3.8$\pm$2.4 \\
\cline{2-5}
&MixedFASNet~\cite{Boulkenafet2017A}&5.3$\pm$6.7&7.8$\pm$5.5&6.5$\pm$4.6 \\
\cline{2-5}
&OFFB~\cite{Wang2018Exploiting}&5.9$\pm$1.9&5.9$\pm$3.0&5.9$\pm$1.0 \\
\cline{2-5}
&STASN~\cite{Yang2019Face}&4.7$\pm$3.9&\textbf{0.9$\pm$1.2}&2.8$\pm$1.6 \\
\cline{2-5}
&Ours&3.6$\pm$1.6&1.4$\pm$2.3&\textbf{2.5$\pm$1.0} \\
\hline
\multirow{5}*{4}
&Auxiliary~\cite{Liu2018Learning}&9.3$\pm$5.6&10.4$\pm$6.0&9.5$\pm$6.0 \\
\cline{2-5}
&GRADIANT~\cite{Boulkenafet2017A}&\textbf{5.0$\pm$4.5}&15.0$\pm$7.1&10.0$\pm$5.0 \\
\cline{2-5}
&OFFB~\cite{Wang2018Exploiting}&14.2$\pm$8.7&\textbf{4.2$\pm$3.8}&9.2$\pm$3.4 \\
\cline{2-5}
&STASN~\cite{Yang2019Face}&6.7$\pm$10.6&8.3$\pm$8.4&\textbf{7.5$\pm$4.7} \\
\cline{2-5}
&Ours&12.7$\pm$7.5&5.2$\pm$9.3&8.9$\pm$3.1 \\
\hline
\end{tabular}
}
\end{center}
\caption{Comparative results of intra test on Oulu-NPU database.}
\label{table2}
\end{table}
\subsection{Cross Testing}
Cross testing aims to justify the generalization potential of the concerned model. We utilize Replay-Attack~\cite{Chingovska2012On} and CASIA-MFSD~\cite{Zhang2012A} which belongs to different distribution to testify the generalization ability of our model. The two databases are split into training and testing part respectively, and we select a training dataset and a testing dataset from different source to train and test. We compare our method with other face anti-spoofing algorithms including tradition solutions such as LBP-TOP~\cite{Pereira2013Can}, Color Texture~\cite{Boulkenafet2017Face} and Spectral cubes~\cite{Allan2015Face}, and some recent excellent methods such as Auxiliary~\cite{Liu2018Learning}, OFFB~\cite{Wang2018Exploiting} and STASN~\cite{Yang2019Face}. Table~\ref{table3} shows the results of cross testing. Our model's performance exceed all state-of-the-art methods by a large margin in crossing testing. It indicates the generalization of our proposed model is the best so far.
\begin{table}[t]
\begin{center}
\begin{tabular}{|c|c|c|c|c|}
\hline
\multirow{3}*{Method}
&Train&Test&Train&Test \\
\cline{2-5}
&CASIA&Replay&Replay&CASIA \\S
&MFSD&Attack&Attack&MFSD \\
\hline
Motion~\cite{Pereira2013Can}&\multicolumn{2}{c|}{50.2$\%$}&\multicolumn{2}{c|}{47.9$\%$} \\
\hline
LBP-TOP~\cite{Pereira2013Can}&\multicolumn{2}{c|}{49.7$\%$}&\multicolumn{2}{c|}{60.6$\%$} \\
\hline
Motion-Mag~\cite{Bharadwaj2013Computationally}&\multicolumn{2}{c|}{50.1$\%$}&\multicolumn{2}{c|}{47.0$\%$} \\
\hline
Spectral cubes~\cite{Allan2015Face}&\multicolumn{2}{c|}{34.4$\%$}&\multicolumn{2}{c|}{50.0$\%$} \\
\hline
LBP~\cite{Boulkenafet2015Face}&\multicolumn{2}{c|}{47.0$\%$}&\multicolumn{2}{c|}{39.6$\%$} \\
\hline
Color Texture~\cite{Boulkenafet2017Face}&\multicolumn{2}{c|}{30.3$\%$}&\multicolumn{2}{c|}{37.7$\%$} \\
\hline
CNN~\cite{Yang2014Learn}&\multicolumn{2}{c|}{48.5$\%$}&\multicolumn{2}{c|}{45.5$\%$} \\
\hline
Auxiliary~\cite{Liu2018Learning}&\multicolumn{2}{c|}{27.6$\%$}&\multicolumn{2}{c|}{28.4$\%$} \\
\hline
FaceDs~\cite{Jourabloo2018Face}&\multicolumn{2}{c|}{28.5$\%$}&\multicolumn{2}{c|}{41.1$\%$} \\
\hline
OFFB~\cite{Wang2018Exploiting}&\multicolumn{2}{c|}{17.5$\%$}&\multicolumn{2}{c|}{24.0$\%$} \\
\hline
STASN~\cite{Yang2019Face}&\multicolumn{2}{c|}{31.5$\%$}&\multicolumn{2}{c|}{30.9$\%$} \\
\hline
ours&\multicolumn{2}{c|}{\textbf{8.0$\%$}}&\multicolumn{2}{c|}{\textbf{22.3$\%$}} \\
\hline
\end{tabular}
\end{center}
\caption{Comparative results of cross test on CASIA-MFSD and Replay-attack database.}
\label{table3}
\end{table}

\subsection{Ablation study}
\textbf{Modules for enhancing feature representation.}
To enhance the ability of capturing spatial and temporal features, we apply feature pyramid modules and pyramid pooling modules for both two branches to fuse features. For each module, we experiment its influence for two branches. Besides, experiments on the direction of feature fusion of feature pyramid module for two branches are also implemented. The protocol 1 of Oulu-NPU and a cross test are utilized to compare comprehensive performance for them. As shown in Table~\ref{table4}, both branches can benefit from feature pyramid and pyramid pooling module. While for the multi-frame branch, a top-down structure for feature fusion is more powerful. In contrast, the RGB difference branch prefers the bottom-up fashion to capture more detailed feature.
\begin{table}[t]
\begin{center}
\begin{tabular}{|c|c|c|c|c|}
\hline
Branch&FPM&PPM&ACER($\%$)&HTER($\%$) \\
\hline
\multirow{4}*{1}
&-&-&5.1&28.3 \\
\cline{2-5}
&top-down&-&8.1&35.5 \\
\cline{2-5}
&bottom-up&-&4.2&21.6 \\
\cline{2-5}
&bottom-up&\checkmark&\textbf{2.0}&\textbf{11.0} \\
\hline
\multirow{4}*{2}
&-&-&8.5&36.0 \\
\cline{2-5}
&bottom-up&-&7.6&24.8 \\
\cline{2-5}
&top-down&-&6.8&20.2 \\
\cline{2-5}
&top-down&\checkmark&\textbf{5.9}&\textbf{16.3} \\
\hline
\end{tabular}
\end{center}
\caption{Ablation study results of feature enhancement modules. Branch 1 is multi-frame based model, and Branch 2 for RGB difference.}
\label{table4}
\end{table}

\textbf{Choice of branches.}
Since our model consists of two sub models which capture different features, we valid the performance of each of them. Besides, we also develop a single-frame RGB version branch that take only one frame as input to valid our power of extracting temporal features, a single origin frame branch that take one full frame without cropping as input to try to increase its ability of distinguish some attacks that can be easily recognized by some background interact information, and an optical flow branch to compare its performance with RGB difference. The single frame branch has the same model structure except for LSTMs, instead the output features are directly concatenated to fed into a fully-connected layer just as the RGB difference branch. The single origin branch is created by a MobileNetV3-Small network without modification, since it aims to learn global context information rather than detailed features such as texture. We extract optical flow by TVL1 algorithm~\cite{Zach2007A} for the optical flow branch which has the same network structure with the RGB difference branch. Each branch are trained individually. After then different combinations of them are experimented to represent multi-dimensional features.

As shown in Table~\ref{table5}, the multi-frame branch performs best in intra testing and cross testing when applying single branch, which verifies its ability of capturing comprehensive features. For single frame based branch, the single frame and single origin frame perform better in different test fashion. The optical flow branch performs poorly compared with the RGB difference. We attribute it to the fact that RGB difference has more detailed information after subtracting two frames directly which may be more suitable for face anti-spoofing. The combination of multi-frame branch and RGB difference branch boosts the performance of intra and cross testing, show their complementary for face anti-spoofing task. Besides, integrating any other branch will not increase accuracy since they may encode redundancy information compared with the two-stream model.

\begin{table}[t]
\begin{center}
\begin{tabular}{|c|c|c|c|c|}
\hline
Branch1&Branch2&Branch3&ACER($\%$)&HTER($\%$) \\
\hline
single&-&-&5.2&28.2 \\
origin&-&-&7.4&16.2 \\
multi&-&-&2.0&11.0 \\
diff&-&-&5.9&16.3 \\
flow&-&-&18.4&35.3 \\
single&multi&-&2.8&22.1 \\
single&diff&-&3.8&20.2 \\
multi&diff&-&\textbf{1.2}&\textbf{8.0} \\
multi&flow&-&5.2&12.1 \\
single&multi&diff&\textbf{1.2}&15.6 \\
origin&multi&diff&3.2&13.6 \\
\hline
\end{tabular}
\end{center}
\caption{Comparative results of combing different branches.}
\label{table5}
\end{table}
\textbf{Exploration Study.}
To better understand and help to develop face anti-spoofing task, we discuss some methods we have tested here. Five improvement fashions that may deemed as useful for similar tasks or original are implemented with our baseline model to enhance generalization: image patch, convLSTM, denoising, style transfer and branch supervise.

Image patch based method take parts of a frame or feature map as inputs, and model spatial-temporal features on each individual part. The final inference are determined by concatenated output features of all parts, which breaks the original distribution to avoid over-fitting and focus on detailed features.

ConvLSTM model temporal features along each spatial kernel, which can be deemed as capture spatial-temporal features immediately~\cite{Shi2015Convolutional}, but it will introduce many extra parameters compared with typical LSTM. For convLSTM based model, we use it to replace the origin LSTM to try to capture more fine-grained spatial-temporal features.

Denoising method is to regard attacks as live face plus noises~\cite{Jourabloo2018Face}. We implement it in our fashion by adding a encoder-decoder module to restore the original input in the front of classification model, and the noise pattern is calculated by subtracting the origin image and the restore image. The final prediction are obtained by a baseline model which take the noise image as input.

In the field of style transfer, gram matrix is widely used to represent style of a image~\cite{Gatys2015Texture}, which is actually texture. And we have known there are differences in texture between genuine and fake faces. As a result, gram matrix of shallow feature maps are computed to contribute to the output feature as complementary in our implement.

Branch supervise is using RGB difference to generate spatial attention masks at different levels respectively. We aim to use these masks to let multi-frame branch focus more on those positions that take changes, therefore, they are then applied in the according layers for multi-frame branch while training it.

Among the five methods, the first two are single-frame based model, the next two take multiple frames as inputs, and the last one is improvement for integrating two stream model. We compare each of them with their baseline model respectively. The backbone network for these experiments are modified MobileNetV3-Small model in our proposed method. The hyper parameters for training and data augment strategy are all as same as above. For each improvement, we have also tested lots of experiments about the chosen of hyper parameters and sub modules, and we only show the results of one typical model for them.

Table~\ref{table6} shows the results of these exploration methods. For improvements of multiple frame branch, the image patch based method incur a loss of accuracy. But we conclude some rules for image patch based methods according to our experiments. First, the image patch base method is useful if we only use output from single layer as final feature representation, as a result we infer multi-layer structure have similar influences of image patch because of multiple scale receptive field. Second, it would be better to divide feature maps in shallower layer than divide image distinctly considering of computing efficiency. The implement of convLSTM also decreases accuracy for face anti-spoofing, which may be due to over-fitting caused by many extra parameters.

The denoising and style transfer methods can both improve performance in cross testing and style transfer method can also boost performance in intra testing at the expense of parameter size. We can conclude that extracting texture features and distribution information simultaneously in the dimension of spatial can bring improvements.

Finally, the branch supervise method is designed to make RGB differences branch help multi-frame branch to increases generalization ability. While the results show that the two stream model have not benefit from this fusion methods. Many experiment results tell us that the independence of two models is important since they may encode complementary and different information, and an over-fitting branch is tend to lead another branch into over-fitting too.

\begin{table}[t]
\begin{center}
\begin{tabular}{|c|c|c|}
\hline
Method&ACER($\%$)&HTER($\%$) \\
\hline
base(multiple)&\textbf{2.0}&\textbf{11.0} \\
image patch&2.8&30.2 \\
convLSTM&3.8&15.3 \\
\hline
base(single)&5.2&28.2 \\
denoising&6.9&25.1 \\
style transfer&\textbf{4.7}&\textbf{27.7} \\
\hline
base(two stream)&\textbf{1.2}&\textbf{8.0} \\
branch supervise&3.4&12.1 \\
\hline
\end{tabular}
\end{center}
\caption{Comparative results of exploration study.}
\label{table6}
\end{table}

\section{Conclusion}
In this paper, we proposed a light and powerful model for face anti-spoofing. Our model consists of two branches, each of them has been proved useful for defending attacks. The multi-frame branches can capture spatial-temporal features and the RGB difference branch is expert in handling detailed motion cues and decrease over-fitting. By applying FPM and PPM feature enhancement modules in both two branches, out model achieves state-of-the-art performance in most intra testing and crossing testing at open databases. While our method can not handle unknown presentation attacks well, which is a disadvantage for it and expected to be solved in the future. We also conduct exploration studies to develop a better understanding of face anti-spoofing task. Some of those improvements can help to increase accuracy under certain conditions, which offer some guidances for future development.

{\small
\bibliographystyle{ieee_fullname}
\bibliography{egbib}
}
\end{document}